\documentclass[runningheads]{llncs}
\usepackage[T1]{fontenc}
\usepackage{graphicx}
\usepackage{xcolor}
\usepackage{subfigure}
\begin{document}

\title{Neural Network Pruning for Real-time Polyp Segmentation}
%
%
\author{Suman Sapkota\inst{1} \and
Pranav Poudel\inst{2} \and
Sudarshan Regmi\inst{1} \and
Bibek Panthi\inst{1} \and
Binod Bhattarai\inst{3}}
\authorrunning{S. Sapkota et al.}
\institute{NepAl Applied Mathematics and Informatics Institute (NAAMII), Nepal \and
Fogsphere (Redev AI Ltd), UK \and
University of Aberdeen, UK}

\maketitle              
\begin{abstract}

Computer-assisted treatment has emerged as a viable application of medical imaging, owing to the efficacy of deep learning models. Real-time inference speed remains a key requirement for such applications to help medical personnel. Even though there generally exists a trade-off between performance and model size, impressive efforts have been made to retain near-original performance by compromising model size. Neural network pruning has emerged as an exciting area that aims to eliminate redundant parameters to make the inference faster. In this study, we show an application of neural network pruning in polyp segmentation. We compute the importance score of convolutional filters and remove the filters having the least scores, which to some value of pruning does not degrade the performance. For computing the importance score we use the Taylor First Order (TaylorFO) approximation of the change in \textit{network output} for the removal of certain filters. Specifically, we employ a gradient-normalized backpropagation for the computation of the importance score. Through experiments in the polyp datasets, we validate that our approach can significantly reduce the parameter count and FLOPs retaining similar performance.

\end{abstract}

\section{Introduction}
Polyp segmentation \cite{fang2019selective,fan2020pranet,mori2022boundary}  is a crucial research problem in the medical domain involving dense classification. The primary aim of segmenting polyps in the colonoscopy and endoscopy is to identify pathological abnormalities in body parts such as the colon, rectum, etc. Such abnormalities can potentially lead to adverse effects causing colorectal cancer, thus inviting fatal damage to health. Statistics show that between 17\% and 28\% of colon polyps are overlooked during normal colonoscopy screening procedures, with 39\% of individuals having at least one polyp missed, according to several recent studies \cite{lee2017risk,kim2017miss}. However, timely diagnosis of a polyp can lead to timely treatment. It has been calculated that a 1\% improvement of polyp detection rate reduces colorectal cancer by 3\% \cite{corley2014adenoma}. Realizing the tremendous upside of early polyp diagnosis, medical AI practitioners have been trying to utilize segmentation models to assist clinical personnel. However, the latency of the large segmentation model has been the prime bottleneck for successful deployment. Utilizing smaller segmentation models is an option, but doing so compromises the performance of the model. In the case of bigger models, there is a good chance model learns significant redundancies thereby leaving room for improvement in performance. In such a scenario, we can prune the parameters of the model to reduce its size for the inference stage. Neural Network Pruning has established itself as an exciting area to reduce the inference time of larger models. 

Neural Network Pruning~\cite{lecun1989optimal,han2015learning,gale2019state,blalock2020state} is one of the methods to reduce the parameters, compute, and memory requirements. This method differs significantly from knowledge distillation~\cite{hinton2015distilling,gou2021knowledge} where a small model is trained to produce the output of a larger model. Neural Network Pruning is performed at multiple levels; (i) weight pruning~\cite{mozer1989using,han2015learning,han2015deep} removes per parameter basis while (ii) neuron/channel~\cite{wen2016learning,lebedev2016fast} pruning removes per neuron or channel basis and (iii) block/group~\cite{gordon2018morphnet,leclerc2018smallify} pruning removes per a block of networks such as residual block or sub-network.  

Weight pruning generally achieves a very high pruning ratio getting similar performance only with a few percentages of the parameters. This allows a high network compression and accelerates the network on specialized hardware and CPUs. However, weight pruning in a defined format such as N:M block-sparse helps in improving the performance on GPUs~\cite{liu2023ten}. Pruning network at the level of neurons or channels helps reduce the parameters with similar performance, however, the pruning ratio is not that high. All these methods can be applied to the same model as well. 



In this work, we are particularly interested in neuron-level pruning. 
Apart from the benefit of reduced parameter, memory, and computation time (or FLOPs), neuron or channel level pruning, the number of neurons in a neural network is small compared to the number of connections and can easily be pruned by measuring the global importance~\cite{lecun1989optimal,hassibi1993optimal,molchanov2016pruning,lee2018snip,yu2018nisp}. We focus on the global importance as it removes the need to inject bias about the number of neurons to prune in each layer. This can simplify our problem to remove less significant neurons globally, allowing us to extend it to differently organized networks such as VGG, ResNet, UNet or any other Architecture. However, in this work, we focus only on the layer-wise, block-wise and hierarchical architecture of UNet~\cite{ronneberger2015u}.

Our experiment on Kvasir Segmentation Dataset using UNet model shows that we can successfully prune $\approx$ 1K Neurons removing $\approx$14\% of parameters and reducing FLOPs requirement to $\approx$ 0.5x the original model with approximately the same performance of the original (from 0.59 IoU to 0.58 IoU). That is half the computational requirement of previous model with negligible performance loss.



\section{Related works}

\subsection{Real-time Polyp Segmentation}
Convolution-based approaches \cite{ronneberger2015u,zhou2018unet++,long2015fully} have mostly dominated the literature while recently attention-based models \cite{fan2020pranet,kim2021uacanet} have also been gaining traction in polyp segmentation.
A number of works have been done in the area of real-time settings too. One of the earliest works \cite{urban2018deep}, evidencing the ability of deep learning models for real-time polyp, has shown to achieve 96\% accuracy in screening colonoscopy. Another work \cite{wang2018development} utilizing a multi-threaded system in a real-time setting, has shown the deep learning models' ability to process at 25 fps with 76.80 ± 5.60 ms latency. Specialized architectures for polyp segmentation have also been studied in the medical imaging literature accounting for real-time performance. MSNet~\cite {zhao2021automatic} introduced a subtraction unit, performing inference on 352x352 at 70 fps, instead of the usual addition as used in many works such as UNet \cite{ronneberger2015u}, UNet++ \cite{zhou2018unet++}, etc. Moreover, NanoNet \cite{jha2021NanoNet} introduced a novel architecture tailor-made for real-time polyp segmentation primarily relying on a lightweight model hence compromising the learning capacity. SANet \cite{wei2021shallow} has been shown to achieve strong performance with an inference speed of about 72 FPS. It showed samples collected under different conditions show inconsistent colors, causing the feature distribution gap and overfitting issue. Another work \cite{qadir2021toward} used 2D gaussian instead of binary maps to better detect flat and small polyps which have unclear boundaries.

\subsection{Neural Network Pruning}


Works in pruning have somewhat lagged behind in medical imaging as compared to other domains. A recent work \cite{bayasi2021culprit} has focused its study on reducing the computational cost of model retraining after post-pruning. 
DNNDeepeningPruning \cite{fernandes2020automatic} proposed the two-stage model development algorithm to build the small model. In the first stage, the residual layers are added until the overfitting starts and in the latter stage, pruning of the model is done with some user instructions. Furthermore, \cite{fernandes2021pruning} has demonstrated evolution strategy-based pruning in generative adversarial networks (GAN) framework for medical imaging diagnostics purposes. In biomedical image segmentation, \cite{jeong2021neural} applied a pruning strategy in U-Net architecture achieving 2x speedup trading off a mere 2\% loss in mIOU(mean Intersection Over Union) on PhC-U373 and DIC-HeLa dataset. STAMP \cite{dinsdale2022stamp} tackles the low data regime through online simultaneous training and pruning achieving better performance with a UNet model of smaller size as compared to the unpruned one. In histological images, the superiority of layer-wise pruning and network-wide magnitude pruning has been shown for smaller and larger compression ratios respectively \cite{mahbod2022deep}. For medical image localization tasks, pruning has also been used to automatically and adaptively identify hard-to-learn examples \cite{jaiswal2023attend}. In our study, we make use of pruning to reduce the model's parameters.



Previous works showed that global importance estimation can be computed using one or all of forward(activation)~\cite{hu2016network}, parameter(weight)~\cite{han2015learning} or backward(gradient)~\cite{wang2020picking,lubana2020gradient,evci2022gradient} signals. Some of the previous techniques use Feature Importance propagation ~\cite{yu2018nisp} or Gradient propagation~\cite{lee2018snip} to find the neuron importance. Others use both activation and gradient information for pruning~\cite{molchanov2016pruning,molchanov2019importance}. Although there are methods using such signals for pruning at initialization~\cite{wang2020picking}, we limit our experiment to the pruning of trained models for a given number of neurons.



In this work, we use importance metric similar to Taylor First Order (Taylor-FO) approximations~\cite{molchanov2016pruning,molchanov2019importance} but from heuristics combining both forward and backward signals. The forward signal, namely the activation of the neuron, and the backward signal, the gradient. We use a normalized gradient signal to make the contribution of each example similar for computing the importance score.



\begin{figure}[t]
    \centering
    \includegraphics[width=\textwidth]{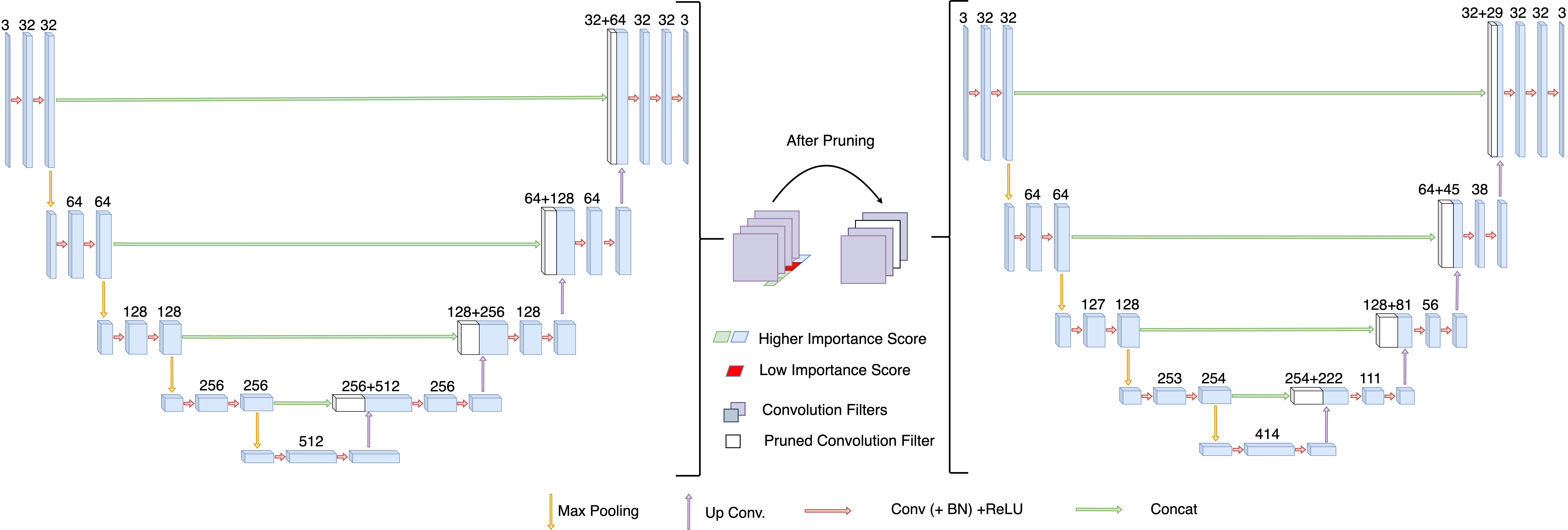}
    \caption{Left: Unpruned UNet Model. Right: Model After Purning convolution filters with low importance score. \textit{The exact number of pruned filters is 956, extracted from experiment shown in Fig~\ref{fig:pruning_iou_flop_param} (top)}.}
    \label{fig:unet_prune}
\end{figure}
\section{Methodology}



In this section, we discuss the pruning method in detail, and the application of the pruning method for polyp segmentation tasks, specifically focusing on the UNet architecture. However, it can be applied to other architecture as well. Instead of pruning all layers, we specifically target the convolutional layers for pruning. It is important to note that the term 'neurons' refers to the channels in the context of pruning convolutional layers. Furthermore, we present a method to select the pruned model that is best suited for the task at hand.

\subsection{Pruning Method}

Previous works on global importance-based post-training pruning of neurons focus on using forward and backward signals. Since most of these methods are based on Taylor approximation of the change in loss after removing a neuron or group of parameters, these methods require input and target value for computing the importance.
Instead, we tackle the problem of pruning from the perspective of overall function output without considering the loss.

\noindent \textbf{Forward Signal:} The forward signal is generally given by the pre-activation ($x_i$). If a pre-activation is zero, then it has no impact on the output of the function, i.e. the output deviation with respect to the removal of the neuron is zero. If the incoming connection of a neuron is zero-weights, then the neuron can be removed, i.e. it has no significance. If the incoming connection is non-zero then the neuron has significance. Forward signal takes into consideration how data affects a particular neuron.\\
\noindent \textbf{Backward Signal:} The backward signal is generally given by back-propagating the loss. If the outgoing connection of the neuron is zeros, then the neuron has no significance to the function, even if it has positive activation. The gradient($\delta x_i$) provides us with information on how the function or loss will change if the neuron is removed. \\
\noindent \textbf{Importance Metric:} Combining the forward and backward signal we can get the influence of the neuron on the loss or the function for given data. Hence, the importance metric ($I_i$) of each neuron ($n_i$) for dataset of size $M$ is given by $I_i = \frac{1}{M}\sum_{n=1}^{M}{x_i.\delta x_i}$, where $x_i$ is the pre-activation and $\delta x_i$ is its gradient. It fulfills the criterion that importance should be low if incoming or outgoing connections are zeros and higher otherwise.\\
\noindent \textit{Problem 1:} This importance metric ($I_i$) is similar to Taylor-FO~\cite{molchanov2016pruning}. However, the metric gives low importance when the gradient is negative, which to our application, is a problem as the function will be changed significantly, even if it lowers the loss. Hence, we calculate the square of importance metric to make it positive.
The squared importance metric ($I_i^{s}$) is computed as below: \\
$$I_i^{s} = \frac{1}{M}\sum_{n=1}^{M}{(x_i.\delta x_i)^2}$$
\noindent \textit{Problem 2:} During the computation of the gradients, some input examples produce a higher magnitude of gradient, and some input examples produce a lower magnitude of the gradient. Since the magnitude is crucial for computing the importance, different inputs contribute differently to the overall importance score. To this end, we normalize the gradient to the same magnitude of $1$. Doing so makes the contribution of each data point equal for computing the importance. \\
\noindent \textbf{Pruning Procedure:}
Consider that pruning is performed using dataset $\mathbf{D} \in [\mathbf{x}_0, \mathbf{x}_1, ... \mathbf{x}_N]$ of size $N$. We have a Convolutional Neural Network (CNN) whose output is given by: $\mathbf{y}_n = f_{CNN}(\mathbf{x}_n)$. We first compute the gradient w.r.t $\mathbf{y}_n$ for all $\mathbf{x}_n$ for given target $\mathbf{t}_n$ as: 
$$\Delta \mathbf{y}_n = \frac{\delta E(\mathbf{y}_n, \mathbf{t}_n)}{\delta \mathbf{y}_n}$$
We then normalize the gradient $\Delta \mathbf{y}_n$ as: 
$$\Delta \mathbf{\hat{y}}_n = \frac{\Delta \mathbf{y}_n}{\|\Delta \mathbf{y}_n\|}$$
This gradient $\Delta \mathbf{\hat{y}}_n$ is then backpropagated through the $f_{CNN}$ network to compute the squared 
Importance score ($I_i^s$) of each convolution filter.
\subsection{Pruning UNet for Polyp-Segmentation}
UNet~\cite{ronneberger2015u} is generally used for Image Segmentation Tasks. It consists of only Convolutional Layers including Upsampling and Downsampling layers organized in a hierarchical structure as shown in Figure~\ref{fig:unet_prune}. 

We compute the Importance Score for each Convolutional layer and prune the least important ones. Removing a single convolution filter removes a channel of the incoming convolution layer and the outgoing convolution channel. When used with many channels, we can get a highly pruned UNet with only a slight change in performance. 

This method can be used to drastically reduce the computation and memory requirements without degrading the performance even without fine-tuning the pruned model. A single computation of Importance Score allows us to prune multiple numbers of neurons and select sparsity with the best FLOPs (or Time-taken) and IoU trade-off. 

\subsection{Measuring Pruning Performance}
Performance metrics are crucial for measuring the effectiveness of different pruning algorithms. Some of them are listed below. \\
\noindent \textbf{FLOPs:}
    FLOP stands for floating point operation. Floating point operation refers to the mathematical operations performed on floating point numbers. FLOP measures model complexity, with a higher value indicating a computationally expensive model and a lower value indicating a computationally cheaper model with faster inference time. We evaluate an algorithm's efficiency by how many FLOPs it reduces. \\ 
\noindent \textbf{Parameters:}
	Parameters represent learnable weights and biases typically represented by floating point numbers. Models with many parameters need a lot of memory, while models with fewer parameters need less memory. The effectiveness of the pruning algorithm is measured by the reduction in the model's parameters. \\
\noindent \textbf{Time-taken:} It is the actual wall-clock inference time of model. We measure time taken before and after pruning the network. Time-taken is practical but not the most reliable metric for efficiency gain as it might vary with device and with different ML frameworks.

\begin{figure*}[ht]
  \centering

\subfigure{\includegraphics[width=0.49\textwidth]{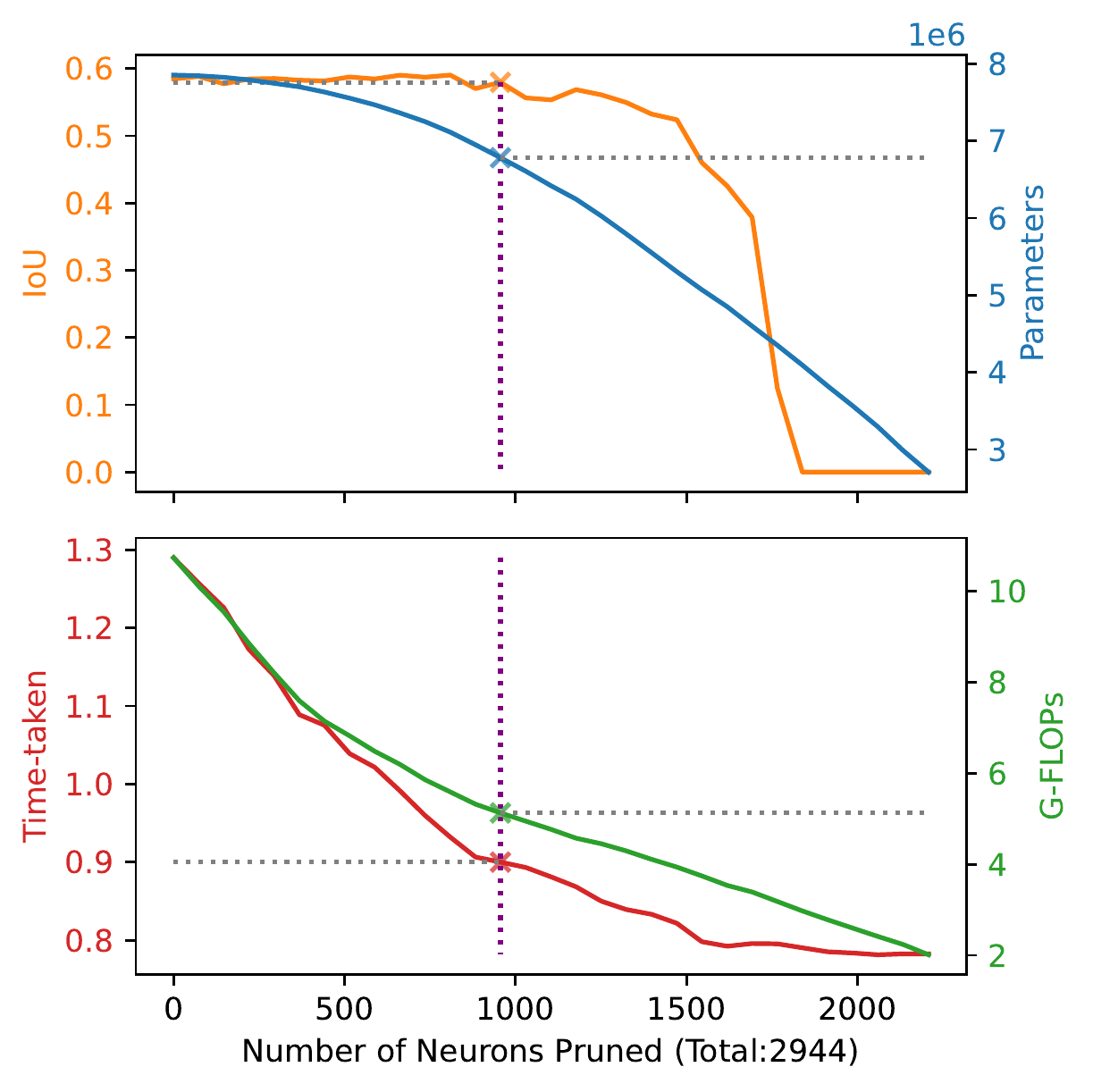}}
\subfigure{\includegraphics[width=0.49\textwidth]{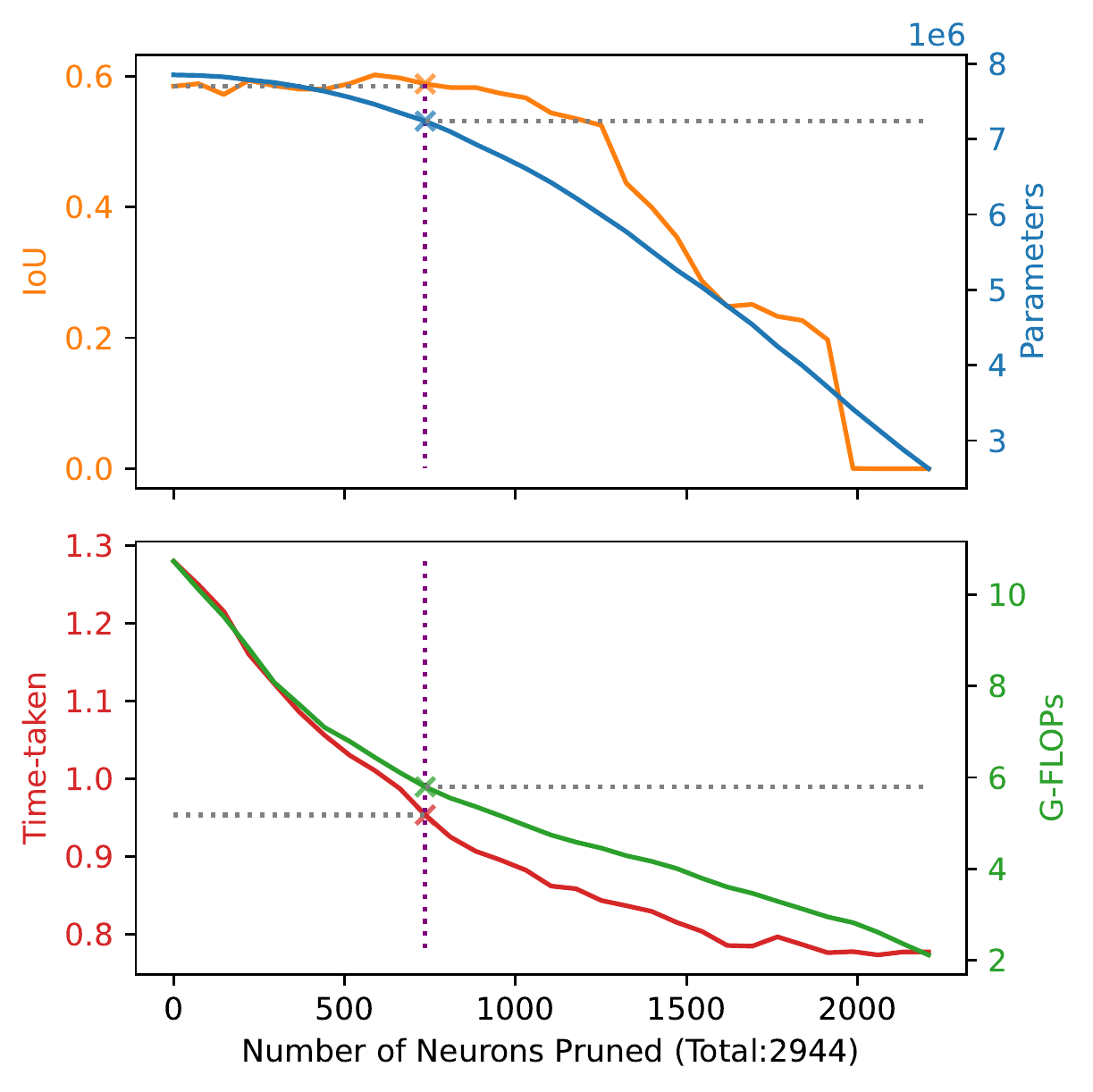}}

\caption{\textbf{(Top)} row is the Number of Neurons Pruned vs IoU and Parameters plot. \textbf{(Bot)} row is the Number of Neurons Pruned vs Time-taken and Giga-FLOPs plot. Here, Time-taken is measured in seconds for the inference of 100 samples with 10 batch size. \textbf{(Left)} column shows pruning performance using 39 data samples for importance estimation. A sample pruning of 956 neurons reduces the FLOPs to 0.477$\times$ and parameters to 0.864$\times$ while retaining performance to 0.99$\times$ the original performance ($\approx$0.5795 IoU). The time taken is reduced by $\approx$30\%.  \textbf{(Right)} column shows pruning performance using 235 data samples for importance estimation. A sample pruning of 736 neurons reduces the FLOPs to 0.54$\times$ and parameters to 0.922$\times$ while retaining the same performance ($\approx$0.5879 IoU). Here, we manage to reduce time taken by $\approx$26\%.}

  \label{fig:pruning_iou_flop_param}
\end{figure*}

\section{Experiments}

\begin{figure*}[ht]
  \centering

\subfigure{\includegraphics[width=0.90\linewidth]{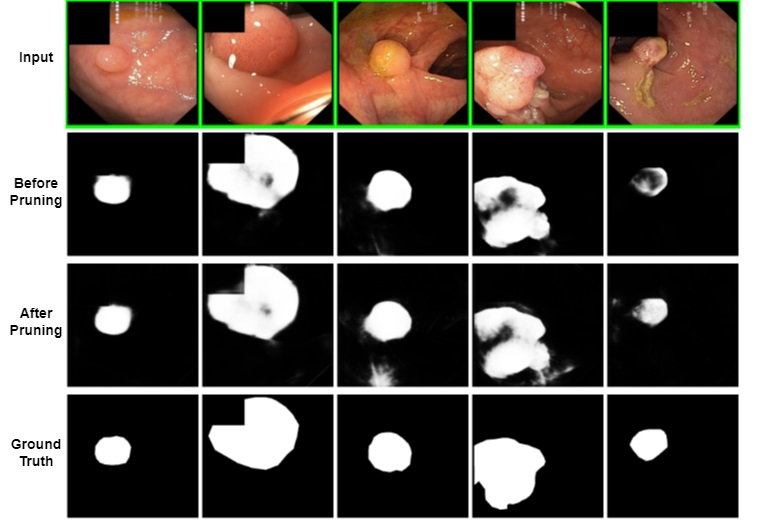}}
  \caption{Qualitative comparison of Polyp Segmentation before and after pruning of the UNet model. The pruned model samples are generated from experiment in Fig~(\ref{fig:pruning_iou_flop_param} \textit{left} with 956 neurons pruned).}
  \label{fig:qualitative_examples_prune}
\end{figure*}


We conduct the experiment for Polyp-Segmentation model pruning using the Kvasir Dataset~\cite{jha2020kvasir}. We use the pretrained UNet Model for segmentation and prune the Convolutional Filters of the Network to reduce the computational cost as shown in Figure~\ref{fig:pruning_iou_flop_param}. \\
\noindent \textbf{Procedure:} First, we compute the importance score for each neuron/channel on a given dataset. Secondly, we prune the $P$ least important neurons of total $N$ by importance metric ($I^s$) given by our method. We measure the resulting accuracy and plot the Number of neurons pruned as shown in Figure~\ref{fig:pruning_iou_flop_param}. The pruning is performed using one split of the test dataset and the IoU is measured on another split of the test dataset.

Although the pruned models could be finetuned to see an increase in performance, we do not finetune pruned model in our case. We analyse the change in performance (IoU), the efficiency achieved (FLOPs) and the compression (the number of parameters) for different values of the \textit{number-of-neurons-pruned} in the UNet Model.\\
\noindent \textbf{Observation:} The experiments show that model generally consists of redundant and less important convolutional channels, which can be pruned with little to no effect on the output of the model. We see in Figure~(\ref{fig:pruning_iou_flop_param} left) that about 50\% ($\approx$1500 out of 2944) of the neurons can be pruned before the IoU starts to decrease drastically. Furthermore, this result is observed for a varying numbers of data points, which suggests that pruning under different settings creates different pruned architectures, while still following the same pattern of performance retention after pruning of an increasing number of neurons (up to some point).

The qualitative evaluation  (see Figure~\ref{fig:qualitative_examples_prune}) of the Pruned UNet Model on the Polyp-Segmentation Dataset shows that the pruned model makes slight changes in the output of the unpruned model while preserving most of the important characteristics. We find that these slight changes can be improvements or degradation to the original model outputs but without significantly distorting the model output.

\section{Conclusion}

In this work, we propose to use Neuron Level Pruning in the application of the polyp segmentation task for the first time. The benefit of proposed channels or filter pruning can be realized immediately with parallel hardware like GPUs to significantly reduce the computation cost to less than 50\% without degrading the performance. Such a reduction in computational cost automatically leads to the potential application in a real-time setting. Computer-assisted treatment of patients, especially during medical tasks like colonoscopy, requires low latency with satisfactory performance such that the pace of treatment is not hindered. Since the polyp's nature can exhibit significant variability during colonoscopy, real-time polyp segmentation models can indeed provide medical personnel useful insights into locating the abnormal growths in the colon thereby assisting in early diagnosis. Moreover, the advanced visualizations aided through real-time diagnosis can indeed lead to determining appropriate treatment approaches. Moreover, it also allows safe, methodical, and consistent diagnosis of patients. Our work paves the path for off-the-shelf models to be significantly accelerated through neural network pruning in tasks requiring fast inference such as medical imaging, reducing the inference and storage cost. To sum up, in this work, we explore a promising research direction of neural network pruning demonstrating its efficacy in polyp segmentation. We validate our approach of neural network pruning with various experiments by almost retaining the original performance.

\section{Acknowledgement}
This work is partly funded by the EndoMapper project by Horizon 2020 FET (GA 863146). 
\bibliographystyle{splncs04}
\bibliography{refs}

\begin{thebibliography}{10}
\providecommand{\url}[1]{\texttt{#1}}
\providecommand{\urlprefix}{URL }
\providecommand{\doi}[1]{https://doi.org/#1}

\bibitem{bayasi2021culprit}
Bayasi, N., Hamarneh, G., Garbi, R.: Culprit-prune-net: Efficient continual
  sequential multi-domain learning with application to skin lesion
  classification. In: Medical Image Computing and Computer Assisted
  Intervention--MICCAI 2021: 24th International Conference, Strasbourg, France,
  September 27--October 1, 2021, Proceedings, Part VII 24. pp. 165--175.
  Springer (2021)

\bibitem{blalock2020state}
Blalock, D., Gonzalez~Ortiz, J.J., Frankle, J., Guttag, J.: What is the state
  of neural network pruning? Proceedings of machine learning and systems
  \textbf{2},  129--146 (2020)

\bibitem{corley2014adenoma}
Corley, D.A., Jensen, C.D., Marks, A.R., Zhao, W.K., Lee, J.K., Doubeni, C.A.,
  Zauber, A.G., de~Boer, J., Fireman, B.H., Schottinger, J.E., et~al.: Adenoma
  detection rate and risk of colorectal cancer and death. New england journal
  of medicine  \textbf{370}(14),  1298--1306 (2014)

\bibitem{dinsdale2022stamp}
Dinsdale, N.K., Jenkinson, M., Namburete, A.I.: Stamp: Simultaneous training
  and model pruning for low data regimes in medical image segmentation. Medical
  Image Analysis  \textbf{81},  102583 (2022)

\bibitem{evci2022gradient}
Evci, U., Ioannou, Y., Keskin, C., Dauphin, Y.: Gradient flow in sparse neural
  networks and how lottery tickets win. In: Proceedings of the AAAI Conference
  on Artificial Intelligence. vol.~36, pp. 6577--6586 (2022)

\bibitem{fan2020pranet}
Fan, D.P., Ji, G.P., Zhou, T., Chen, G., Fu, H., Shen, J., Shao, L.: {PraNet}:
  Parallel reverse attention network for polyp segmentation. In: Medical Image
  Computing and Computer Assisted Intervention (MICCAI) (2020)

\bibitem{fang2019selective}
Fang, Y., Chen, C., Yuan, Y., Tong, K.y.: Selective feature aggregation network
  with area-boundary constraints for polyp segmentation. In: Medical Image
  Computing and Computer Assisted Intervention--MICCAI 2019: 22nd International
  Conference, Shenzhen, China, October 13--17, 2019, Proceedings, Part I 22.
  pp. 302--310. Springer (2019)

\bibitem{fernandes2020automatic}
Fernandes, F.E., Yen, G.G.: Automatic searching and pruning of deep neural
  networks for medical imaging diagnostic. IEEE Transactions on Neural Networks
  and Learning Systems  \textbf{32}(12),  5664--5674 (2020)

\bibitem{fernandes2021pruning}
Fernandes~Jr, F.E., Yen, G.G.: Pruning of generative adversarial neural
  networks for medical imaging diagnostics with evolution strategy. Information
  Sciences  \textbf{558},  91--102 (2021)

\bibitem{gale2019state}
Gale, T., Elsen, E., Hooker, S.: The state of sparsity in deep neural networks.
  arXiv preprint arXiv:1902.09574  (2019)

\bibitem{gordon2018morphnet}
Gordon, A., Eban, E., Nachum, O., Chen, B., Wu, H., Yang, T.J., Choi, E.:
  Morphnet: Fast \& simple resource-constrained structure learning of deep
  networks. In: Proceedings of the IEEE conference on computer vision and
  pattern recognition. pp. 1586--1595 (2018)

\bibitem{gou2021knowledge}
Gou, J., Yu, B., Maybank, S.J., Tao, D.: Knowledge distillation: A survey.
  International Journal of Computer Vision  \textbf{129},  1789--1819 (2021)

\bibitem{han2015deep}
Han, S., Mao, H., Dally, W.J.: Deep compression: Compressing deep neural
  networks with pruning, trained quantization and huffman coding. arXiv
  preprint arXiv:1510.00149  (2015)

\bibitem{han2015learning}
Han, S., Pool, J., Tran, J., Dally, W.: Learning both weights and connections
  for efficient neural network. Advances in neural information processing
  systems  \textbf{28} (2015)

\bibitem{hassibi1993optimal}
Hassibi, B., Stork, D., Wolff, G.: Optimal brain surgeon: Extensions and
  performance comparisons. Advances in neural information processing systems
  \textbf{6} (1993)

\bibitem{hinton2015distilling}
Hinton, G., Vinyals, O., Dean, J.: Distilling the knowledge in a neural
  network. arXiv preprint arXiv:1503.02531  (2015)

\bibitem{hu2016network}
Hu, H., Peng, R., Tai, Y.W., Tang, C.K.: Network trimming: A data-driven neuron
  pruning approach towards efficient deep architectures. arXiv preprint
  arXiv:1607.03250  (2016)

\bibitem{jaiswal2023attend}
Jaiswal, A., Chen, T., Rousseau, J.F., Peng, Y., Ding, Y., Wang, Z.: Attend who
  is weak: Pruning-assisted medical image localization under sophisticated and
  implicit imbalances. In: Proceedings of the IEEE/CVF Winter Conference on
  Applications of Computer Vision. pp. 4987--4996 (2023)

\bibitem{jeong2021neural}
Jeong, T., Bollavaram, M., Delaye, E., Sirasao, A.: Neural network pruning for
  biomedical image segmentation. In: Medical Imaging 2021: Image-Guided
  Procedures, Robotic Interventions, and Modeling. vol. 11598, pp. 415--425.
  SPIE (2021)

\bibitem{jha2020kvasir}
Jha, D., Smedsrud, P.H., Riegler, M.A., Halvorsen, P., de~Lange, T., Johansen,
  D., Johansen, H.D.: Kvasir-{SEG}: A segmented polyp dataset. In: MultiMedia
  Modeling: 26th International Conference, MMM 2020, Daejeon, South Korea,
  January 5--8, 2020, Proceedings, Part II 26. pp. 451--462. Springer (2020)

\bibitem{jha2021NanoNet}
Jha, D., Tomar, N.K., Ali, S., Riegler, M.A., Johansen, H.D., Johansen, D.,
  de~Lange, T., Halvorsen, P.: Nanonet: Real-time polyp segmentation in video
  capsule endoscopy and colonoscopy. In: 2021 IEEE 34th International Symposium
  on Computer-Based Medical Systems (CBMS). pp. 37--43 (2021).
  \doi{10.1109/CBMS52027.2021.00014}

\bibitem{kim2017miss}
Kim, N.H., Jung, Y.S., Jeong, W.S., Yang, H.J., Park, S.K., Choi, K., Park,
  D.I.: Miss rate of colorectal neoplastic polyps and risk factors for missed
  polyps in consecutive colonoscopies. Intestinal research  \textbf{15}(3),
  ~411 (2017)

\bibitem{kim2021uacanet}
Kim, T., Lee, H., Kim, D.: {UACANet}: Uncertainty augmented context attention
  for polyp segmentation. In: Proceedings of the 29th ACM International
  Conference on Multimedia. pp. 2167--2175 (2021)

\bibitem{lebedev2016fast}
Lebedev, V., Lempitsky, V.: Fast convnets using group-wise brain damage. In:
  Proceedings of the IEEE conference on computer vision and pattern
  recognition. pp. 2554--2564 (2016)

\bibitem{leclerc2018smallify}
Leclerc, G., Vartak, M., Fernandez, R.C., Kraska, T., Madden, S.: Smallify:
  Learning network size while training. arXiv preprint arXiv:1806.03723  (2018)

\bibitem{lecun1989optimal}
LeCun, Y., Denker, J., Solla, S.: Optimal brain damage. Advances in neural
  information processing systems  \textbf{2} (1989)

\bibitem{lee2017risk}
Lee, J., Park, S.W., Kim, Y.S., Lee, K.J., Sung, H., Song, P.H., Yoon, W.J.,
  Moon, J.S.: Risk factors of missed colorectal lesions after colonoscopy.
  Medicine  \textbf{96}(27) (2017)

\bibitem{lee2018snip}
Lee, N., Ajanthan, T., Torr, P.H.: Snip: Single-shot network pruning based on
  connection sensitivity. arXiv preprint arXiv:1810.02340  (2018)

\bibitem{liu2023ten}
Liu, S., Wang, Z.: Ten lessons we have learned in the new" sparseland": A short
  handbook for sparse neural network researchers. arXiv preprint
  arXiv:2302.02596  (2023)

\bibitem{long2015fully}
Long, J., Shelhamer, E., Darrell, T.: Fully convolutional networks for semantic
  segmentation. In: Proceedings of the IEEE Conference on Computer Vision and
  Pattern Recognition. pp. 3431--3440 (2015)

\bibitem{lubana2020gradient}
Lubana, E.S., Dick, R.P.: A gradient flow framework for analyzing network
  pruning. arXiv preprint arXiv:2009.11839  (2020)

\bibitem{mahbod2022deep}
Mahbod, A., Entezari, R., Ellinger, I., Saukh, O.: Deep neural network pruning
  for nuclei instance segmentation in hematoxylin and eosin-stained
  histological images. In: Applications of Medical Artificial Intelligence:
  First International Workshop, AMAI 2022, Held in Conjunction with MICCAI
  2022, Singapore, September 18, 2022, Proceedings. pp. 108--117. Springer
  (2022)

\bibitem{molchanov2019importance}
Molchanov, P., Mallya, A., Tyree, S., Frosio, I., Kautz, J.: Importance
  estimation for neural network pruning. In: Proceedings of the IEEE/CVF
  conference on computer vision and pattern recognition. pp. 11264--11272
  (2019)

\bibitem{molchanov2016pruning}
Molchanov, P., Tyree, S., Karras, T., Aila, T., Kautz, J.: Pruning
  convolutional neural networks for resource efficient inference. arXiv
  preprint arXiv:1611.06440  (2016)

\bibitem{mozer1989using}
Mozer, M.C., Smolensky, P.: Using relevance to reduce network size
  automatically. Connection Science  \textbf{1}(1),  3--16 (1989)

\bibitem{qadir2021toward}
Qadir, H.A., Shin, Y., Solhusvik, J., Bergsland, J., Aabakken, L., Balasingham,
  I.: Toward real-time polyp detection using fully cnns for 2d gaussian shapes
  prediction. Medical Image Analysis  \textbf{68},  101897 (2021)

\bibitem{mori2022boundary}
Qiu, J., Hayashi, Y., Oda, M., Kitasaka, T., Mori, K.: Boundary-aware feature
  and prediction refinement for polyp segmentation. Computer Methods in
  Biomechanics and Biomedical Engineering: Imaging \& Visualization pp. 1--10
  (2022)

\bibitem{ronneberger2015u}
Ronneberger, O., Fischer, P., Brox, T.: U-net: Convolutional networks for
  biomedical image segmentation. In: Medical Image Computing and
  Computer-Assisted Intervention--MICCAI 2015: 18th International Conference,
  Munich, Germany, October 5-9, 2015, Proceedings, Part III 18. pp. 234--241.
  Springer (2015)

\bibitem{urban2018deep}
Urban, G., Tripathi, P., Alkayali, T., Mittal, M., Jalali, F., Karnes, W.,
  Baldi, P.: Deep learning localizes and identifies polyps in real time with
  96\% accuracy in screening colonoscopy. Gastroenterology  \textbf{155}(4),
  1069--1078 (2018)

\bibitem{wang2020picking}
Wang, C., Zhang, G., Grosse, R.: Picking winning tickets before training by
  preserving gradient flow. arXiv preprint arXiv:2002.07376  (2020)

\bibitem{wang2018development}
Wang, P., Xiao, X., Glissen~Brown, J.R., Berzin, T.M., Tu, M., Xiong, F., Hu,
  X., Liu, P., Song, Y., Zhang, D., et~al.: Development and validation of a
  deep-learning algorithm for the detection of polyps during colonoscopy.
  Nature biomedical engineering  \textbf{2}(10),  741--748 (2018)

\bibitem{wei2021shallow}
Wei, J., Hu, Y., Zhang, R., Li, Z., Zhou, S.K., Cui, S.: Shallow attention
  network for polyp segmentation. In: Medical Image Computing and Computer
  Assisted Intervention--MICCAI 2021: 24th International Conference,
  Strasbourg, France, September 27--October 1, 2021, Proceedings, Part I 24.
  pp. 699--708. Springer (2021)

\bibitem{wen2016learning}
Wen, W., Wu, C., Wang, Y., Chen, Y., Li, H.: Learning structured sparsity in
  deep neural networks. Advances in neural information processing systems
  \textbf{29} (2016)

\bibitem{yu2018nisp}
Yu, R., Li, A., Chen, C.F., Lai, J.H., Morariu, V.I., Han, X., Gao, M., Lin,
  C.Y., Davis, L.S.: Nisp: Pruning networks using neuron importance score
  propagation. In: Proceedings of the IEEE conference on computer vision and
  pattern recognition. pp. 9194--9203 (2018)

\bibitem{zhao2021automatic}
Zhao, X., Zhang, L., Lu, H.: Automatic polyp segmentation via multi-scale
  subtraction network. In: Medical Image Computing and Computer Assisted
  Intervention--MICCAI 2021: 24th International Conference, Strasbourg, France,
  September 27--October 1, 2021, Proceedings, Part I 24. pp. 120--130. Springer
  (2021)

\bibitem{zhou2018unet++}
Zhou, Z., Rahman~Siddiquee, M.M., Tajbakhsh, N., Liang, J.: Unet++: A nested
  u-net architecture for medical image segmentation. In: Deep Learning in
  Medical Image Analysis and Multimodal Learning for Clinical Decision Support:
  4th International Workshop, DLMIA 2018, and 8th International Workshop,
  ML-CDS 2018, Held in Conjunction with MICCAI 2018, Granada, Spain, September
  20, 2018, Proceedings 4. pp. 3--11. Springer (2018)

\end{thebibliography}
\end{document}